\documentclass[pamm,a4paper,fleqn]{w-art}
\usepackage{times,cite,w-thm}
\usepackage[T1]{fontenc}
\usepackage[utf8]{inputenc}
\usepackage{multicol}       
\usepackage{graphicx}

\newcommand{\be}{\begin{equation}}
\newcommand{\ee}{\end{equation}}
\newcommand{\bc}{\begin{center}}
\newcommand{\ec}{\end{center}}
\newcommand{\bd}{\begin{description}}
\newcommand{\ed}{\end{description}}
\newcommand{\bi}{\begin{itemize}}
\newcommand{\ei}{\end{itemize}}

\newcommand{\bs}{\boldsymbol}

\newcommand{\bmat}{\begin{pmatrix}}
\newcommand{\emat}{\end{pmatrix}}
\newcommand{\bsmat}{\left(\begin{smallmatrix}}
\newcommand{\esmat}{\end{smallmatrix}\right)}
\newcommand{\bes}{\begin{equation}\begin{split}}
\newcommand{\ees}{\end{split}\end{equation}}

\begin{document}

\TitleLanguage[EN]
\title[]{A data-driven model order reduction approach for Stokes flow through random porous media}

\author{\firstname{Constantin} \lastname{Grigo}\inst{1,}%
\footnote{Corresponding author: e-mail \ElectronicMail{constantin.grigo@tum.de}, 
     phone  +49 (89) 289 - 15218}} 
\address[\inst{1}]{\CountryCode[DE] Department of Mechanical Engineering,
Technical University of Munich,
Boltzmannstr.\ 15,
85748 Garching}
\author{\firstname{Phaedon-Stelios} \lastname{Koutsourelakis}\inst{1,}%
     \footnote{e-mail p.s.koutsourelakis@tum.de}}
\AbstractLanguage[EN]
\begin{abstract}
Direct numerical simulation of Stokes flow through an impermeable, rigid body matrix by finite elements requires meshes fine enough to resolve the pore-size scale and is thus a computationally expensive task. The cost is significantly  amplified when randomness in the pore microstructure is present and therefore multiple simulations need to be carried out. It is well known that in the limit of scale-separation, Stokes flow can be accurately  approximated by Darcy’s law \cite{Whitaker1986} with an effective diffusivity field depending on viscosity and the pore-matrix  topology.  We propose a fully probabilistic, Darcy-type, reduced-order model which, based on only a few tens of full-order Stokes model runs, is capable of  learning a map from the fine-scale topology to the effective diffusivity and  is maximally predictive of the  fine-scale response. The reduced-order model learned can significantly accelerate uncertainty quantification tasks as well as provide quantitative confidence metrics of the predictive estimates produced.  
\end{abstract}
\maketitle                   

\vspace{-10mm}
\section{Introduction}
Numerical simulations of physical or engineering systems are  hampered by the presence of high-dimensional uncertainties in  their input parameters, such as external loads, material- or geometric properties. Monte Carlo estimators,  while dimension-independent, can lead to a large amount of forward model evaluations which are not computationally affordable in many cases.

A prominent approach to reduce the computational cost is by replacing the forward model by a much cheaper, yet inaccurate surrogate or emulator trained on only a small number of forward runs. Machine learning methods such as Gaussian Processes or Deep Neural Networks have grown into prominence but suffer from the curse of dimensionality and/or require an excessively large amount of data to be trained.

In this work, we introduce a fully Bayesian surrogate to Stokes flow through random porous media, which is utilizing a simplified physics emulator (Darcy flow) as its core unit and is therefore capable of dealing with large-dimensional input uncertainties (thousands) whilst only requiring few (tens) forward model runs for training.

\vspace{-.35cm}
\section{Methodology}

\subsection{The full-order model: Stokes flow through random porous media}
\begin{figure}[h]
    \centering
    \includegraphics[width=.78\textwidth]{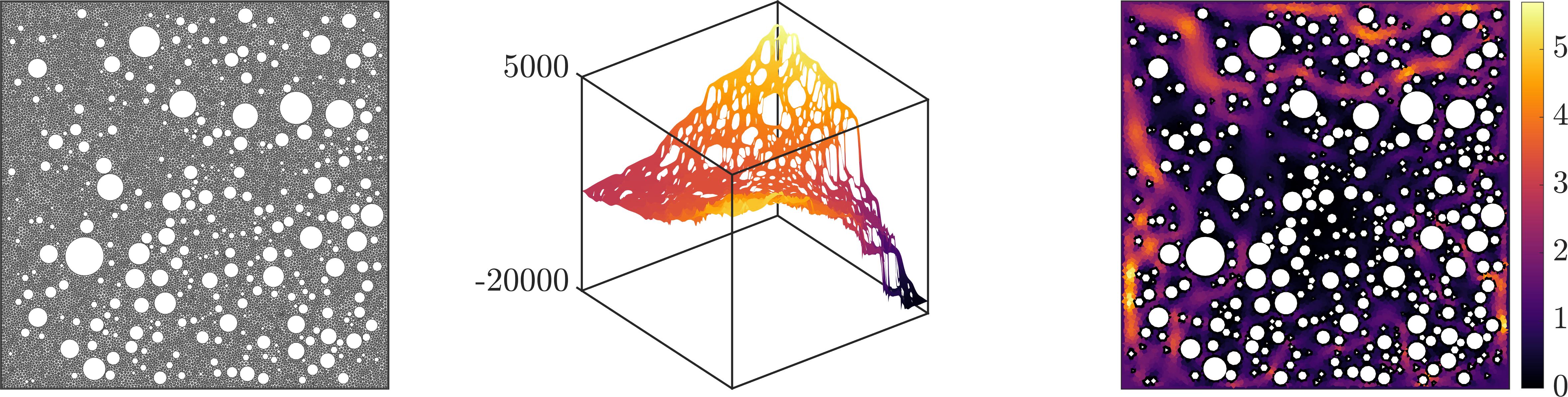}
    \caption{FE mesh (left), pressure response $p$ (middle) and norm of velocity field $|\bs v|$ for Stokes flow through a typical porous medium realization with non-overlapping polydisperse circular exclusions.}
    \label{data}
\end{figure}
\vspace{-.35cm}
We consider porous microstructures described as perforated unit square domains with impermeable inclusions. Assuming unit viscosity $\mu = 1$, the strong form of the fully resolved Stokes flow problem is given by \newline
\begin{minipage}{.49\linewidth}
\begin{subequations}
\begin{align}
    -\bs \nabla \cdot \bs \nabla \bs v + \bs \nabla P &= 0 & &\text{in} \quad \Omega_F \\
    \bs \nabla \cdot \bs v &= 0 & &\text{in} \quad \Omega_F
\end{align}
\end{subequations}
\end{minipage}
\hfill
\begin{minipage}{.49\linewidth}
\setcounter{equation}{0}
\begin{subequations}
\setcounter{equation}{2}
\begin{align}
    \bs t &= \bs n \cdot (\bs \nabla \bs v - \bs I P) = -P_0\bs n & &\text{on} \quad \Gamma_p \\
    \bs v &= \bs v_0 & &\text{on} \quad \Gamma_{\bs v} \\
    \bs v &= \bs 0  & &\text{on} \quad \Gamma_{\text{int}}
\end{align}
\end{subequations}
\end{minipage}
where $\Omega_F$ is the pore domain, $\Gamma_p, \Gamma_{\bs v}, \Gamma_{\text{int}}$ denote external Neumann, Dirichlet and interfacial (Dirichlet) boundaries and $\bs v$, $P$, and $\bs t$ are the flow velocity, pressure field and Cauchy tractions, respectively. We use Taylor-Hood elements and call $\bs U_f = \bsmat \bs P_f \\ \bs V_f\esmat$ the FE solution to the pressure and velocity fields $p$ and $\bs v$ evaluated on a regular $128\times 128$ grid.

\vspace{-.35cm}

\subsection{A fully Bayesian reduced-order model}
We propose a fully Bayesian, three-component reduced-order model \cite{Grigo2017a}, \cite{Grigo2017b} comprising:
\begin{itemize}
    \item a probabilistic map from the high-dimensional vector $\bs \lambda_f$ characterizing the random microstructure/domain (e.g. a list of centers and radii of circular exclusions) to a much lower dimensional vector $\bs \lambda_c$ representing an effective diffusivity random field; this probabilistic mapping is denoted with $p_c(\bs \lambda_c|\bs \lambda_f, \bs \theta_c)$, where $\bs\theta_c$ parametrized by $\bs \theta_c$;
    \item a deterministic finite element solver for Darcy's equation,
    \begin{equation}
        -\bs \nabla \cdot(\lambda_c \bs \nabla P_c) = 0 
    \end{equation}
    mimicking Stokes flow on a much coarser scale; We call $\bs P_c = \bs P_c(\bs \lambda_c)$ the Darcy FE pressure response and $\bs V_c = \bs V_c(\bs \lambda_c)$ the corresponding flux field, both evaluated on the same grid as the fine scale data;
    \item a probabilistic coarse-to-fine mapping from $\bs U_c = \bsmat \bs P_c(\bs \lambda_c) \\ \bs V_c(\bs \lambda_c) \esmat$ to the fine scale response $\bs U_f = \bsmat \bs P_f \\ \bs V_f\esmat$. We call this mapping $p_{cf}(\bs U_f|\bs U_c, \bs \theta_{cf})$, parametrized by $\bs \theta_{cf}$;
\end{itemize}

We pursue a fully Bayesian approach and marginalize all model parameters $\bs \theta_c, \bs \theta_{cf}$ to get the predictive distribution
\begin{equation}
p(\bs U_f|\bs \lambda_f) = \int p_{cf}(\bs U_f|\bs U_c(\bs \lambda_c), \bs \theta_{cf}) p_c(\bs \lambda_c|\bs \lambda_f, \bs \theta_c) p_{\bs \theta_{cf}}(\bs \theta_{cf}) p_{\bs \theta_c}(\bs \theta_c) d\bs \lambda_c d\bs \theta_c d\bs \theta_{cf},
\end{equation}
where the effective Darcy diffusivity $\bs \lambda_c$ appears as a latent variable. Model training is performed using an adapted form of the Variational Relevance Vector Machine \cite{Bishop2000} suitable for latent variable models.

\vspace{-.35cm}

\section{Example}
\begin{figure}[h]
    \centering
    \includegraphics[width=.78\textwidth]{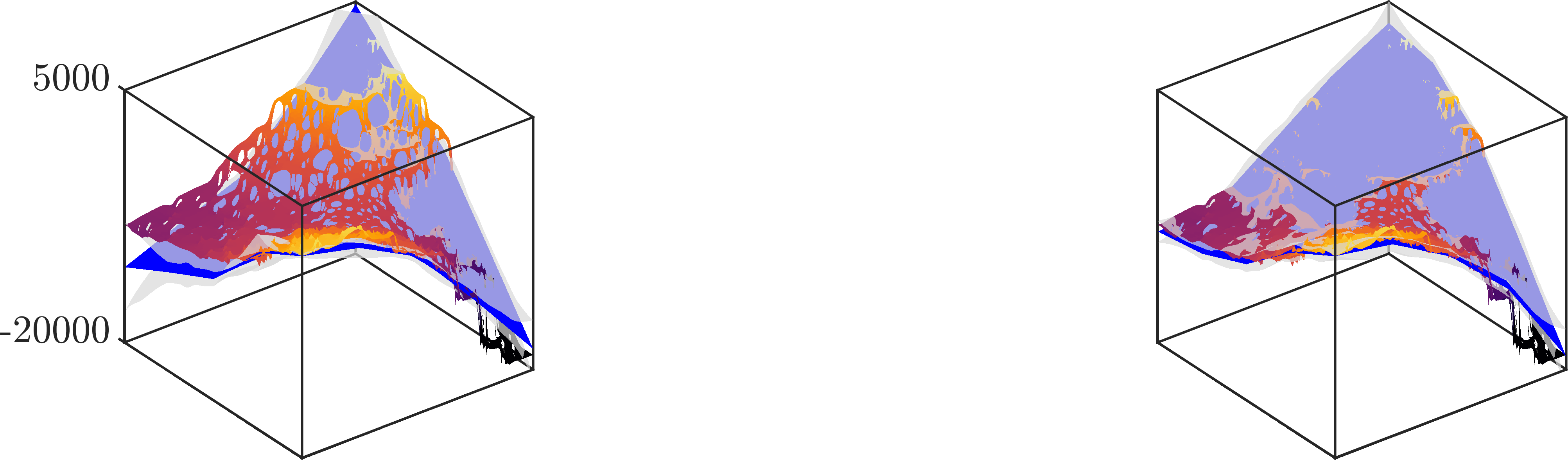}
    \caption{Predictive examples on the Stokes flow pressure response $p$ for $N = 8$ training samples and Darcy emulator resolution $2\times 2$ (left) and $N = 80$ using a $4\times 4$ emulator mesh (right). The blue surface is the predictive mean $\pm$ 1 standard deviation (transparent grey). The colored surface is the true response of a test sample.}
    \label{predictions}
\end{figure}
For training/testing our model, we use microstructures with totally impenetrable polydisperse spherical exclusions (see also Figure \ref{data}). The number and radii of exclusions are sampled from a (discretized) $\log$-normal distribution. Predictive examples for low/high number of training data and different Darcy emulator discretizations are shown in Figure \ref{predictions}.

\vspace{-.5cm}
\section{Conclusion}
We have introduced a fully Bayesian, three-component probabilistic surrogate model for Stokes flow through random porous media using a much cheaper model emulator based on Darcy flow on a much coarser spatial discretization as the key component. Model evaluations on test samples show tight distributions centered around the true solution. Predictive errors improve with the number of training data and with the resolution of the Darcy-based emulator.

\vspace{\baselineskip}

\bibliographystyle{pamm}
\bibliography{references}

\end{document}